\documentclass[10pt,twocolumn,letterpaper]{article}

\usepackage{cvpr}              

\usepackage[dvipsnames]{xcolor}

\definecolor{cvprblue}{rgb}{0.21,0.49,0.74}
\usepackage[pagebackref,breaklinks,colorlinks,citecolor=cvprblue]{hyperref}

\usepackage{bbm}
\usepackage{multirow}
\usepackage[symbol]{footmisc}

\newcommand{\jinzhi}[1]{{\color{black}#1}}
\newcommand{\haiyang}[1]{{\color{black}#1}}
\newcommand{\yixuan}[1]{{\color{black}#1}}
\def\paperID{4354} 
\def\confName{CVPR}
\def\confYear{2024}

\title{OmniSeg3D: Omniversal 3D Segmentation via \\ Hierarchical Contrastive Learning}

\author{Haiyang Ying$^1$,\ \ 
        Yixuan Yin$^1$,\ \ 
        Jinzhi Zhang$^1$,\ \ 
        Fan Wang$^2$,\ \ 
        Tao Yu$^1$,\ \ 
        Ruqi Huang$^1$,\ \ 
        Lu Fang$^{1\dagger}$
        \vspace{0.1cm}
    \and
        $^1$Tsinghua University, \ \ 
        $^2$Alibaba Group 
}%

\begin{document}

\twocolumn[{%
\renewcommand\twocolumn[1][]{#1}%
\maketitle
\thispagestyle{empty}
\begin{center}
    \centering
    \captionsetup{type=figure}
    \includegraphics[width=1.0\textwidth]{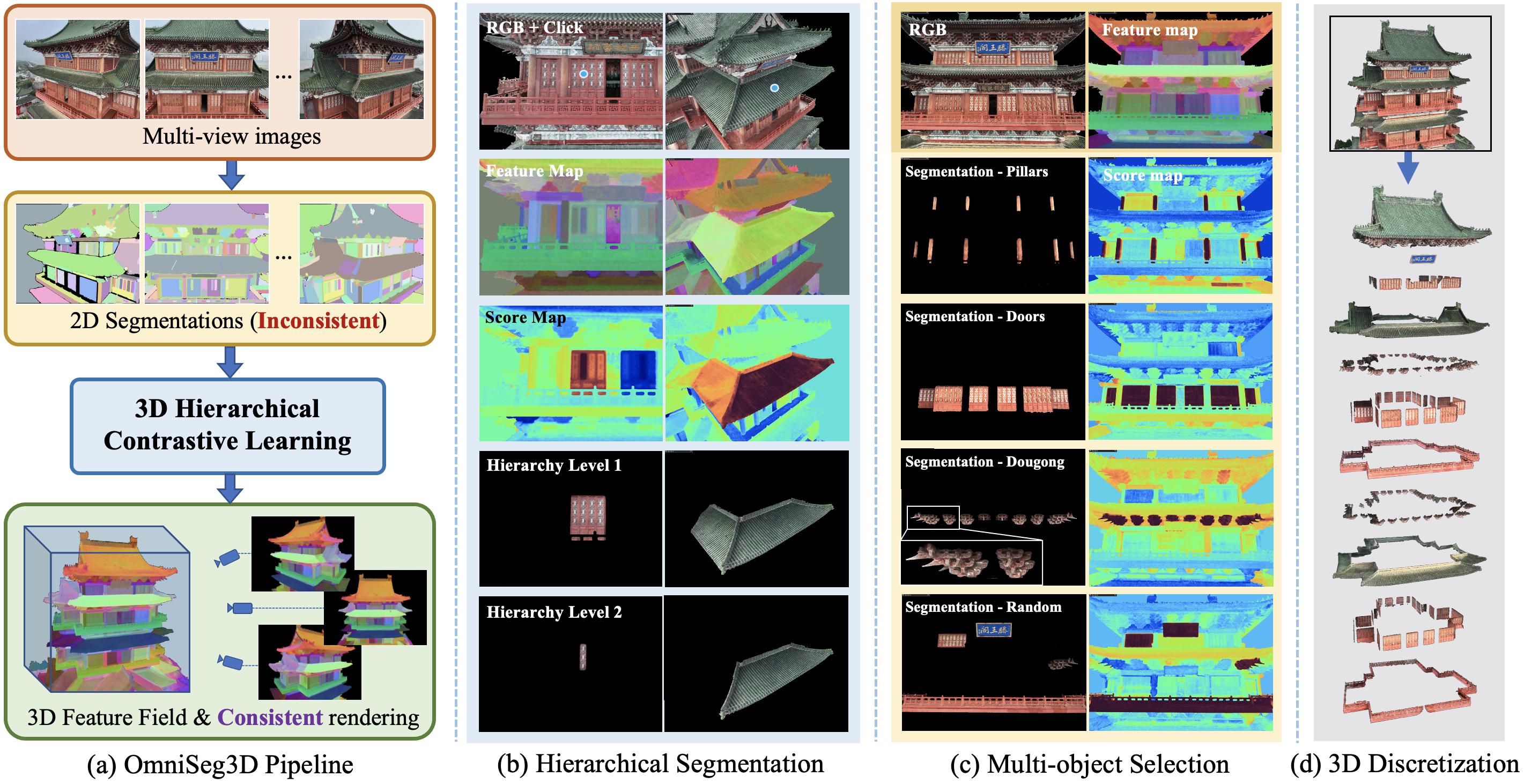}
    \captionof{figure}{
    We propose an omniversal 3D segmentation method, which (a) takes as input multi-view, inconsistent, and class-agnostic 2D segmentations,  and outputs a consistent 3D feature field via a hierarchical contrastive learning framework. This method supports (b) hierarchical segmentation, (c) multi-object selection, and (d) holistic discretization in an interactive manner. 
    \href{https://oceanying.github.io/OmniSeg3D/}{Project Page.}
    }
    \label{fig.teaser}
\end{center}%
}]

\footnotetext[2]{Corresponding author 
}

\newcommand{\V}{\mathbf}
\newcommand{\norm}[1]{\Vert#1\Vert}

\begin{abstract}
   Towards holistic understanding of 3D scenes, a general 3D segmentation method is needed that can segment diverse objects without restrictions on object quantity or categories, while also reflecting the inherent hierarchical structure.
   To achieve this, we propose OmniSeg3D, an omniversal segmentation method aims for segmenting anything in 3D all at once. 
   The key insight is to lift multi-view inconsistent 2D segmentations into a consistent 3D feature field through a hierarchical contrastive learning framework, which is accomplished by two steps. 
   Firstly, we design a novel hierarchical representation based on category-agnostic 2D segmentations to model the multi-level relationship among pixels. 
   Secondly, image features rendered from the 3D feature field \jinzhi{are} clustered at different levels, which \jinzhi{can be} further drawn closer or \yixuan{pushed} apart according to the hierarchical relationship between different levels.
   \yixuan{In tackling} the challenges posed by inconsistent 2D segmentations, this framework yields a global consistent 3D feature field, which further enables hierarchical segmentation, multi-object selection, and global discretization.
   Extensive experiments demonstrate the effectiveness of our method on high-quality 3D segmentation and accurate hierarchical structure understanding.
   A graphical user interface further facilitates flexible interaction for omniversal 3D segmentation.    
\end{abstract}

\section{Introduction}
3D segmentation forms one of the cornerstones in 3D scene understanding, which is also the basis of 3D interaction, editing, and extensive applications in virtual reality, medical analysis, and robot navigation. 
To meet the requirement of complex world sensing, a general/omniversal category-agnostic 3D scene segmentation method is required, capable of segmenting any object in 3D without limitations on object quantity or categories. For instance, to accurately discretize a pavilion as shown in Fig.~\ref{fig.teaser}, the user needs to accurately segment each roof, column, eaves, and other intricate structures.
Existing 3D-based segmentation methods based on 3D point clouds, meshes, or volumes fall short of these requirements. They are either restricted to limited categories due to the scarcity of large-scale 3D datasets, such as learning-based methods~\cite{han2020occuseg, liu2022insconv, kirillov2019panopticseg}, or they only identify local geometric similarity or smoothness without extracting semantic information, as typified by traditional algorithms~\cite{schnabel2007efficientransac, hohne1992trad3Dseg1, dorninger20073d}.

An alternative approach involves lifting 2D image understanding to 3D space, leveraging the impressive class-agnostic 2D segmentation performance achieved by recent methods~\cite{kirillov2023sam, li2022lseg, chen2022focalclick, liu2023simpleclick, sofiiuk2022ritm}.
%
Current lifting-based methods either rely on annotated 2D masks~\cite{zhi2021semanticnerf, bing2022dmnerf, wu2022objectsdf}, or are restricted to a limited set of pre-defined classes~\cite{siddiqui2023panopticlift, bhalgat2023contrastivelift}. Other methods propose distilling semantic-rich image features~\cite{radford2021clip, li2022lseg} onto point clouds~\cite{peng2023openscene, takmaz2023openmask3d} or NeRF~\cite{kerr2023lerf, kobayashi2022dff, goel2023isrf}. 
However, due to the absence of boundary information, directly distilling these semantic feature into 3D space often leads to noisy segmentations~\cite{peng2023openscene, kerr2023lerf}.
Further works use SAM~\cite{kirillov2023sam} or video segmentation methods~\cite{mirzaei2023spinnerf} to generate accurate 2D masks of \yixuan{targeted} objects, and unproject them into 3D space~\cite{cen2023samin3d}. 
However, these approaches are limited to single-object segmentation and exhibit unstable results in cases with severe occlusion because the 2D segmentation is performed on each image independently.

Therefore, significant challenges still persist. First, multi-view consistency remains an obstacle due to the substantial variations in 2D segmentations across different viewpoints. Second, ambiguity arises when distinguishing in-the-wild objects like eaves and roofs, which inherently possess a hierarchical semantic structure.
To this end, we propose OmniSeg3D, an \textbf{Omniversal 3D Segmentation} method which enjoys multi-object, category-agnostic, and hierarchical segmentation in 3D all at once.
We demonstrate that a global 3D feature field (which can be formulated on point cloud, mesh, NeRF~\cite{mildenhall2021nerf}, etc) is inherently well-suited for integrating occlusion-free, boundary-clear, and hierarchical semantic information from 2D segmentations through hierarchical contrastive learning.
%
The key lies in hierarchically clustering 2D image features rendered from the 3D feature field at different levels of segmentation blocks, where the multi-level segmentations are specified by a proposed hierarchical 2D representation. Then the clustered features will be drawn closer or pushed apart via a hierarchical contrastive loss, which enables the learning of a feature field that encodes hierarchical information into the proximity of feature distances, effectively eliminating semantic inconsistencies between different images.
%
This unified framework facilitates multi-object selection, hierarchical segmentation, global discretization, and a broad range of applications.


We evaluate OmniSeg3D on segmentation tasks for single object selection and hierarchical inference. Extensive quantitative and qualitative results on real-world and synthetic datasets demonstrate our method enjoys high-quality 3D object segmentation and holistic comprehension of scene structure across various scales. An interactive interface is also provided for flexible 3D segmentation.
Our contributions are summarized as follows:
\begin{itemize}
    \item We propose a \textbf{hierarchical 2D representation} to reveal and store the part-level relationship within objects based on class-agnostic 2D segmentations and a voting strategy.
    \item We present a \textbf{hierarchical contrastive learning method} to optimize a globally consistent 3D hierarchical feature field given 2D observations.
    \item Extensive experiments demonstrate that our \textbf{omniversal 3D segmentation framework} can segment anything in 3D all at once, which enables hierarchical segmentation, multi-object selection, and 3D discretization. 
\end{itemize}

\section{Related Works}
\subsection{2D Segmentation}
2D segmentation has experienced a long history.
Early works mainly rely on the clue of pixel similarity and continuity~\cite{felzenszwalb2004efficient, coleman1979image, achanta2012slic} to segment images.
Since the introduction of FCN~\cite{long2015fcn}, there has been a rapid expansion in research of different sub-fields of 2D segmentation~\cite{he2017maskrcnn, kirillov2019panopticseg, chen2017deeplab, zhao2017pyramid}. The involvement of transformer~\cite{vaswani2017attention} in the segmentation domain has led to the proposal of several novel segmentation architectures~\cite{zheng2021rethinking, cheng2021perpixel, cheng2022maskedattention}. However, most of these methods are limited to pre-defined class labels.

Prompt-based segmentation is a special task that enables segmenting unseen object categories~\cite{liu2023simpleclick, chen2022focalclick, sofiiuk2022ritm}.
One recent breakthrough is the Segment Anything Model (SAM)~\cite{kirillov2023sam}, aiming to unify the 2D segmentation task through the introduction of a prompt-based segmentation approach , is considered a promising innovation in the field of vision.


\subsection{3D Segmentation}

\noindent{\textbf{Closed-set segmentation.}} 
The task of 3D segmentation has been explored with various types of 3D representation such as RGBD images~\cite{wang2018depth, xing2020malleable}, pointcloud~\cite{yi2019gspn, yang20193dbonet, hu2020randlanet}, and voxels~\cite{huang2016point, liu2019point, han2020occuseg, liu2022insconv}. However, due to the insufficiency of annotated 3D datasets for training a unified 3D segmentation model, they are still limited to closed-set 3D understanding, which largely restrict the application scenarios.

\begin{figure*}[thp]
  \centering
   \includegraphics[width=0.9\linewidth]{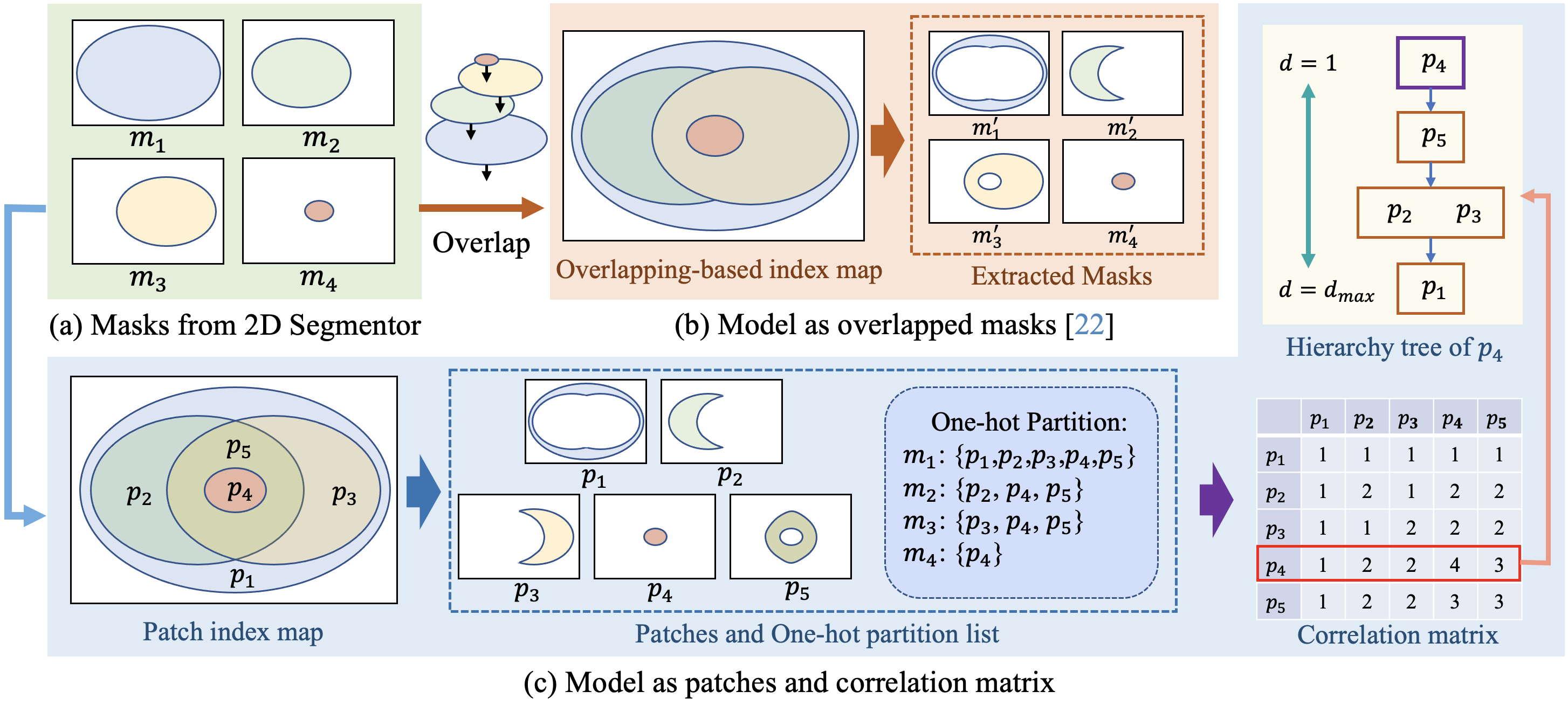}
   \caption{Illustration of our proposed hierarchical representation. (a) For each image, click-based 2D segmentors provide a set of masks $\{m_i\}$. (b) \jinzhi{Directly overlapping masks implemented by conventional methods}~\cite{kirillov2023sam} lead to the loss of hierarchical information. (c) 
   Patch-based modeling effectively preserves inclusion information. 
   The hierarchical representation of each image includes a patch index map $I_p$ and a correlation matrix $C_{hi}$, where the relevance between $p_i$ and other patches is evaluated via a voting strategy.
   }
   \label{fig:representation}
\end{figure*}

Given the shortage of 3D datasets essential for the development of foundational 3D models,  recent works have proposed to lift 2D information into 3D for 3D segmentation and understanding.
Some works rely on ground truth masks~\cite{zhi2021semanticnerf, bing2022dmnerf, wu2022objectsdf} or pre-trained 2D semantic/instance segmentation models for mask generation~\cite{siddiqui2023panopticlift, bhalgat2023contrastivelift}. However, ground truth annotation is unrealistic for general cases, and model-based methods typically only offer closed-set object masks. 
ContrastiveLift~\cite{bhalgat2023contrastivelift} proposes to segment closed-set 3D objects via contrastive learning. However, it cannot handle unseen classes and reveal object hierarchy. In contrast, our method achieves panoptic, category-agnostic, and hierarchical segmentation based on a hierarchical contrastive learning framework, which can be interpreted as a sound combination of click-based segmentation methods and holistic 3D modeling.

\vspace{6pt}
\noindent{\textbf{Open-set segmentation.}} LERF~\cite{kerr2023lerf} and subsequent works~\cite{goel2023isrf, kobayashi2022dff, takmaz2023openmask3d} propose to distill language feature~\cite{radford2021clip} into 3D space for open-vocabulary interactive segmentation. Since the learned feature is trained on entire images without explicit boundary supervision, these methods prone to produce noisy segmentation boundaries. Besides, these methods are unable to distinguish different instances due to the lack of instance-level supervision. 
Alternatively, we take advantage of category-agnostic segmentation methods and distill the 2D results into 3D to get a consistent feature field and enable high-quality 3D segmentation.

SPInNeRF~\cite{mirzaei2023spinnerf} utilizes video segmentation to initialize 2D masks and then lift them into 3D space with a NeRF. A followed multi-view refinement stage is utilized to achieve consistent 3D segmentation. 
%
SA3D~\cite{cen2023samin3d} introduces an online interactive segmentation method that propagates one SAM~\cite{kirillov2023sam} mask into 3D space and other views iteratively. However, these methods may heavily rely on a good choice of reference view and cannot handle complex cases such as severe occlusion.
Instead, our method can segment anything in 3D all at once via a global consistent feature field, which is more robust to object occlusion.

\vspace{6pt}
\noindent{\textbf{Hierarchical segmentation.}}
For hierarchical segmentation, existing methods mainly focus on category-specific scenario~\cite{mo2019partnet, mo2019structurenet} or geometric analysis~\cite{chen2020bspnet, yu2022caprinet, yi2019gspn}, which are not suitable for general hierarchical 3D segmentation. Instead, we propose to distill hierarchical information from 2D into 3D space to achieve multi-view consistent hierarchical segmentation in 3D.

\section{Methods}


Given a set of calibrated input images and \jinzhi{the corresponding} \haiyang{2D segmentation masks}, our goal is to learn a 3D feature field that enjoys multi-object, category-agnostic, and hierarchical segmentation all at once.
We first segment 2D images into smaller units \haiyang{$P_{segs}$} and construct our novel hierarchical 2D representation. 
Then we hierarchically cluster 2D image features \haiyang{$\mathbf{f} \in \mathbb{R}^D$} rendered from the 3D feature field at different levels of patches \haiyang{$P_{segs}$}, which will further be supervised to construct correct feature distance order between sampled points via the proposed hierarchical contrastive clustering strategy.
In this section, we first introduce the representation in Sec.~\ref{sec:representation}, which includes both basic and hierarchical implementation for lifting inconsistent 2D masks into 3D space. Then, a hierarchical contrastive learning method for optimizing the 3D feature field will be discussed (Sec.~\ref{sec:HCC}). Finally, the applications for various interactive segmentation will be introduced.


\subsection{Hierarchical Representation}\label{sec:representation}

\textbf{Preliminary: Class-agnostic 2D segmentation.}
To achieve omniversal segmentation, a 2D segmentation method should be able to handle unseen categories. We seek solution from click-based method like SAM~\cite{kirillov2023sam}, which exhibits a class-agnostic property.
Given an input image $I$, a grid of points (typically $32\times32$) are sampled as the input prompts to generate a set of 2D binary masks $M_{segs}=\{ m_i\in \mathbb{R}^{H\times W} | i=1,...,|M_{segs}| \}$ as proposals (see Fig.~\ref{fig:representation}(a)).
%
To get a label map as training data for 3D field optimization (like in \cite{zhi2021semanticnerf}), masks in $M_{segs}$ are overlapped one by one according to the number of contained pixels in the masks in \cite{kirillov2023sam} (see Fig.~\ref{fig:representation}(b)).
Since each pixel in image $I$ may belong to more than one masks in $M_{segs}$ (consider the fact that a pixel belonging to the mask of a chair may also belong to the mask of the chair's leg), directly overlapping masks, as done in SAM, may destroy the rich hierarchical information embedded inside $M_{segs}$.

\vspace{6pt}
\noindent\textbf{Hierarchical Modeling.}
To avoid the aforementioned problem, we design a novel representation that preserves the hierarchical information within each image.
%
Specifically, instead of using overlapped masks, we divide the entire 2D image into disjoint patches. As shown in Fig.~\ref{fig:representation}(a), let $m_i\in M_{segs}, (i=1,...,4)$ represent masks in $M_{segs}$. 
For each pixel, we create a one-hot vector to indicate which masks the pixel belongs to.
To eliminate the impact of overlapping, we define the patch set $P_{segs}$ as the smallest collection of pixels that share the same one-hot vector. 
These patches can also be interpreted as the smallest units in the image that are exhaustively partitioned by $M_{segs}$ (as shown in Fig.~\ref{fig:representation}(c)). This also results in a patch index map $I_p$, where each pixel contains a index of the patch.

Next, we proceed to model the hierarchical structure with patches $P_{segs}$ as the unit and the original masks $M_{segs}$ as the \jinzhi{correlation binding}. The core idea is that, if two patches fall into the same mask, then these two patches has some degree of correlation. To model the strength of the correlation, we introduce a voting-based rating strategy. 
Specifically, for each pair of patches $p_i$ and $p_j$, we count the number of masks that contain both $p_i$ and $p_j$. By traversing all the patch pairs, we get a matrix $C_{hi}\in \mathbb{R}^{N_p \times N_p}$:
\begin{equation}
    C_{hi}(p_i, p_j) = \sum_{k=1}^{N_m} \mathbbm{1}(p_i \subseteq m_k) \cdot \mathbbm{1}(p_j \subseteq m_k),
\end{equation}
where $N_m = |M_{segs}|$ represents the number of masks and $N_p = |P_{segs}|$ represents the number of patches. $N_p$ typically ranges from $200$ to $500$ in our experiments.
This process can be interpreted as utilizing masks to vote for the relationship between patches.
To deduce the hierarchical relationship between patches, we select a patch $p_i$ as the anchor and take the $i$-th row of matrix $C_{hi}(p_i, \cdot)=v_i$.
We then sort the patches according to the vote counts in vector $v_i$ and construct a hierarchical tree for anchor patch $p_i$, as illustrated in Fig.~\ref{fig:representation}(c).
Patches located at shallower depths in the tree has stronger relevance to the anchor patch $p_i$, which can be taken as the guidance of the hierarchical contrastive learning introduced in the subsequent section.
Finally, the hierarchical representation for each image consists of a patch index map $I_p$ and a correlation matrix $C_{hi}$.

%




\begin{figure*}[t] 
\centering
    \includegraphics[width=0.9\linewidth]{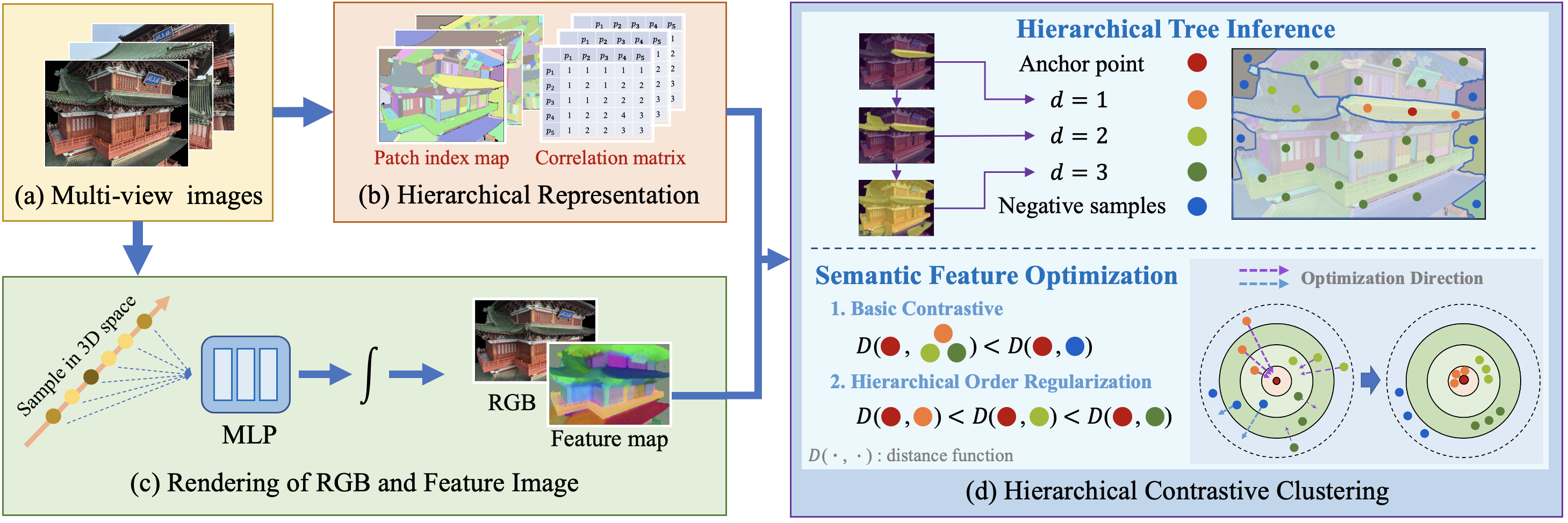} 
    \caption{Framework of hierarchical contrastive learning. (a) For each input RGB image, we apply (b) 2D hierarchical modeling to get a \jinzhi{patch index map} and a correlation matrix. During training, we utilize (c) \jinzhi{NeRF-based} rendering pipeline to render features from 3D space and apply hierarchical contrastive learning (d) to the rendered features to optimize the feature field for segmentation.}\label{fig:framework}
\end{figure*}

\subsection{Hierarchical Contrastive Learning} \label{sec:HCC}
In this section, we show how to lift the hierarchical relationship of 2D patches into the 3D space through hierarchical contrastive learning.


\noindent{\textbf{3D feature field. }}
We start by introducing a 3D feature field that establishes the relationship between 2D images and the 3D space. This feature field is based on NeRF-like rendering methods~\cite{mildenhall2021nerf, muller2022instantngp}. Specifically, for each point $\mathbf{x}_i \in \mathbb{R}^3$ in the 3D space, we define a segmentation identity feature \jinzhi{$\mathbf{f_i} \in \mathbb{R}^D$}.
Along the view direction $\mathbf{d}_i \in \mathbb{R}^2$, an MLP network $F_{\mathbf{\Theta}}$ generates per-point attributes:
\begin{equation}
    (\sigma_i, \mathbf{f}_i) = F_{\mathbf{\Theta}}(\gamma_1(\mathbf{x}_i)), 
    \ \ \ 
    \mathbf{c}_i = F_{\mathbf{\Theta}}(\gamma_1(\mathbf{x}_i), \gamma_2(\mathbf{d}_i)),
    \label{eq:nerf_mlp}
\end{equation}
where $\gamma_1$ and $\gamma_2$ are positional encoding functions in \cite{muller2022instantngp}.

Subsequently, color and density are integrated along the ray to generate the rendered pixel color $\mathbf{c}(\mathbf{r})$:
\begin{equation}
    \mathbf{c}(\mathbf{r})=\sum_{i=1}^{N} T_i\alpha_i \mathbf{c}_i, 
    \ \ \ 
    T_i=\prod_{j=1}^{i-1} (1-\alpha_i),
    \label{eq:volume_rendering}
\end{equation}
where $\alpha_i = 1-exp(-\sigma_i\delta_i)$ is the opacity and $\delta_i= r_{i+1} - r_i$ is the distance between adjacent samples. Besides, feature maps can be rendered as:
\begin{equation}
    \mathbf{f}(\mathbf{r})=\sum_{i=1}^{N} T_i\alpha_i \mathbf{f}_i . 
    \label{eq:volume_rendering_2}
\end{equation}

\noindent{\textbf{Basic implementation.}}
\label{sec:basic}
{In this section, we present a basic implementation} of our approach that lifts 2D segmentations into 3D space without considering hierarchical information.

{The core idea is} to apply contrastive learning to lift 2D category-agnostic segmentation to 3D.
For each image, we randomly sample $N$ points on it and identify the patch id each point belongs to. Then we render features $\{ \mathbf{f_i} \} (i\in [1, N])$ of these points via differentiable rendering \jinzhi{from the 3D feature field}.
For each sampled point, we designate points with the same patch id as positive samples, and all the other sampled points as negative ones. The correlation between two 3D points is modelled as the cosine distance $\mathbf{f}_i \cdot \mathbf{f}_j$.

To accelerate the loss calculation and get stable convergence, we apply the contrastive clustering method~\cite{li2020pcl}. Specifically, we define cluster $\{\mathbf{f}^i\}$ as the collection of rendered features that share the same patch id $i$. The center of each cluster is defined as the mean value $\bar{\mathbf{f}}^i$ of features in $\{\mathbf{f}^i\}$. Then for each chosen feature point $\mathbf{f}^i_j$ with patch id $i$ and point index $j$ within cluster $\{\mathbf{f}^i\}$, both positive samples and negative samples are replaced with the mean feature $\bar{\mathbf{f}}^i$ and $\bar{\mathbf{f}}^k$. 
The loss is shown below, {which favors high similarity between $\mathbf{f}^i_j$ and $\bar{\mathbf{f}}^i$ \jinzhi{that belongs to the same cluster} and low similarity between $\mathbf{f}^i_j$ and $\bar{\mathbf{f}}^k$}:
\begin{equation}
\mathcal{L}_{CC}=-\frac{1}{N_p}\sum_{i=1}^{N_p}\sum_{j=1}^{|\{\mathbf{f}^i\}|}\log{\frac{\exp(\mathbf{f}^{i}_{j}\cdot \bar{\mathbf{f}}^i/ \phi_i)}{\sum_{k=1}^{N_p}\exp(\mathbf{f}^{i}_{j}\cdot \bar{\mathbf{f}}^k/ \phi_k)}},
\label{eq:cc}
\end{equation}
where $N_p$ is the number of patch ids, $\phi_i$ is the temperature of cluster $i$ to balance the cluster size and variance:
$\phi_i={\sum_{j=1}^{n_i}||\mathbf{f}^i_j-\bar{\mathbf{f}}^i||_2} / {n_i\log(n_i+\alpha)},  \ \ n_i=|\{\mathbf{f}^i\}|$.
$\alpha=10$ is a {smoothing} parameter to prevent small clusters from exhibiting an excessively large $\phi_i$.

Note that ConstrastiveLift~\cite{bhalgat2023contrastivelift} uses a slow-fast learning strategy for stable training. We refer to contrastive clustering~\cite{li2020pcl} to realize faster training and stable convergence.

\vspace{6pt}
\noindent{\textbf{Hierarchical implementation.}}
Here we show how to incorporate hierarchical information into the pipeline of contrastive learning. 
%
\jinzhi{We first cluster the sampled point features ${\mathbf{f}}$ into feature point sets $\{\mathbf{f}^i\}, (i\in [1, N_p])$ based on the 2D image patches. }
Then for each anchor patch ${p_i}$, we assign all related patches with their depths in the hierarchy tree $d\in [1, d^i_{max}]$ according to the \jinzhi{correlation} matrix $C_{hi}$.
Note that all the related patches are potential positive samples in this formulation.

To achieve hierarchical contrastive clustering in 3D, we employ the hierarchical regularization proposed in~\cite{zhang2022useallthelabels}. 
Firstly, we add a regularization term $\lambda^{d-1}$ to Eq.~\ref{eq:cc} with a per-level decay factor $\lambda \leq 1$, 
\haiyang{which means higher penalty are applied to the patches with stronger correlations to the anchor patch $i$}.
Secondly, a regularization of the optimization order is implemented to ensure that a patch higher in the hierarchy tree \haiyang{(smaller $d$)} exhibits a higher \haiyang{feature similarity} with the anchor patch than patches at lower levels (as shown in Fig.~\ref{fig:framework}(d)). \jinzhi{The final loss is shown below}:
%
%
\begin{equation} \mathcal{L}_{H}=\sum_{i=1}^{N_p}\sum_{d=1}^{d^i_{max}} \frac{\lambda^{d-1}}{NL}\sum_{j=1}^{|\{\mathbf{f}^i\}|} \sum_{s\in S_d^i}  \max(\mathcal{L}^{i,j}(s),\mathcal{L}_{max}^{i,j}(d-1)) ,
\label{eq:reg}
\end{equation}
where 
{$S^i_d$ is the \jinzhi{patch index \haiyang{set}} at level $d$ of anchor patch $i$ (For example, $S_{d=3}^{i=4}=\{2, 3\}$ in Fig.~\ref{fig:representation}), $ s\in S^i_d$ is a patch at depth $d$, }
$\mathcal{L}^{i,j}(s)$ is the contrastive loss between point $j$ (in point set of patch $i$) and the average feature $\bar{\mathbf{f}}^s$ of patch $s$:
\begin{equation}
\mathcal{L}^{i,j}(s)=
-\log{\frac{\exp(\mathbf{f}^{i}_{j}\cdot \bar{\mathbf{f}}^s/ \phi_s)}{\sum_{k=1}^{N_p}\exp(\mathbf{f}^{i}_{j}\cdot \bar{\mathbf{f}}^k/ \phi_k)}} , 
\end{equation}
\jinzhi{and} $\mathcal{L}_{max}^{i,j}(d)$ is the maximum \jinzhi{loss at level $d$}:
\begin{equation}
    \mathcal{L}_{max}^{i,j}(d)=\max_{s\in S^i_d}\mathcal{L}^{i,j}(s).
\end{equation}
%
Since the volumetric rendering may introduce ambiguity in the calculation of the integration, we found that it is better to apply normalization loss to regularize the feature vector and ensure it distributed on the sphere surface:
\begin{equation}
\mathcal{L}_{norm}=\frac{1}{N}\sum_{i=1}^{N}\left({||\mathbf{f}_i||-1}\right)^2 .
\end{equation}

\begin{figure*}[t] 
\centering
    \includegraphics[width=0.9\linewidth]{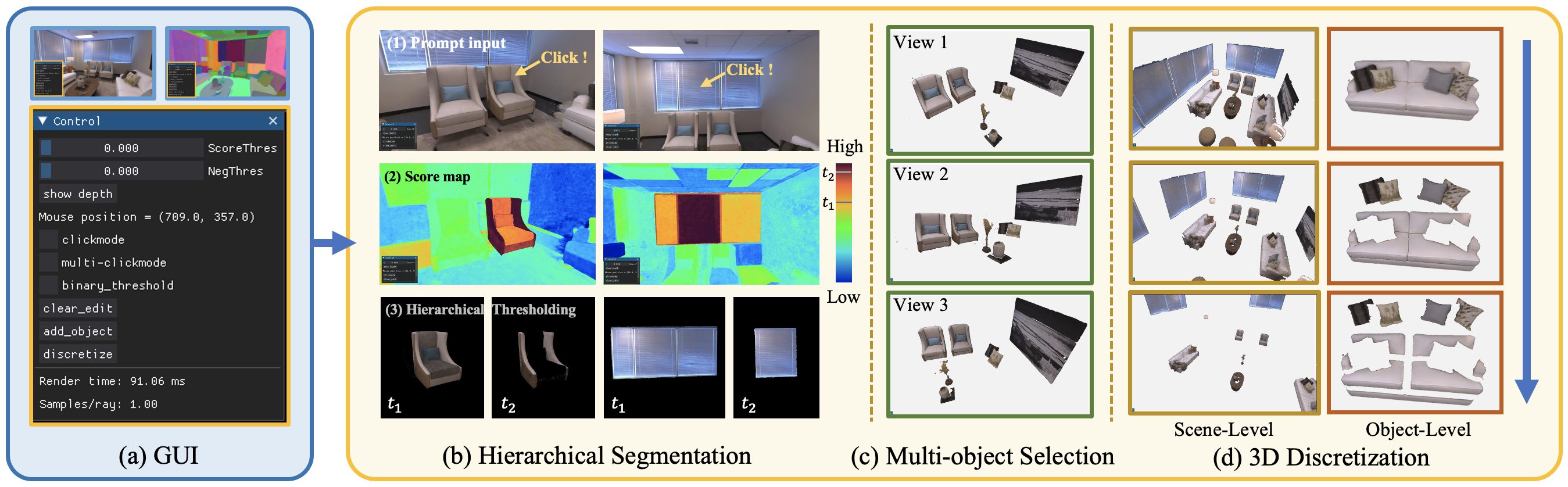} 
    \caption{Interactive 3D segmentation with (a) a graphical user interface. \haiyang{For \texttt{room-0} of Replica, we show the segmentation performance on (b) hierarchical inference, (c) multi-object selection, and (d) 3D discretization with our GUI.}}
    \label{fig:gui}
\end{figure*}

\subsection{Implementation details}

During training, we optimize the MLP $F_\mathbf{\Theta}$ and the semantic feature volume $\mathbf{V_s}$ (with feature dimension $D=16$) via volume rendering, where four loss functions are applied: 
$\mathcal{L}_c = \sum_{\mathbf{r}}\left \| \mathbf{c}(\mathbf{r})-\mathbf{c}_{\mathrm{gt}}(\mathbf{r}) \right \|^2_2$, 
%
%
$\mathcal{L}_{reg} = \sum_{\mathbf{r}} -o(\mathbf{r})\log(o(\mathbf{r}))$,
where $o(\mathbf{r})=\sum_{i=1}^{N} T_i\alpha_i$ is the opacity of each ray. $\mathcal{L}_{reg}$ is used to regularize each ray to be completely saturated or empty. The per-level decay factor is set to $\lambda=0.5$. The total loss is: 
\begin{equation}
    \mathcal{L}_{total}= \mathcal{L}_{c} + w_1\mathcal{L}_{H} + w_2\mathcal{L}_{norm} + w_3\mathcal{L}_{reg}
\end{equation}
The hyper-parameters are set to $w_1 = 5\times 10^{-4}, w_2=5\times 10^{2}, w_3=1\times 10^{-3}$ for all the experiments. 
With a cosine annealing schedule, the learning rate is set from $1\times 10^{-2}$ to $3\times 10^{-4}$. The number of rays in each batch is 8192. We train our model for 50000 iterations for each scene. 

The proposed omniversal segmentation scheme can be seen as a lightweight plug-in which can be easily integrated into reconstruction methods based on common 3D representations like NeRF, mesh, and point cloud. 
For 2D backbones, though we use SAM~\cite{kirillov2023sam} in our implementation, any click-based segmentation methods like ~\cite{sofiiuk2022ritm, liu2023simpleclick, chen2022focalclick} can be used as a substitute.
Please refer to our supplementary material for more details.

\subsection{Interactive Segmentation}
To realize flexible and interactive 3D segmentation, we develop a graphical user interface (GUI).
This GUI can serve as a novel 3D annotation tool, which may largely improve the efficiency of 3D data annotation. Two typical cases based on NeRF and mesh are shown in Fig.~\ref{fig:gui} and Fig.~\ref{fig.teaser}.

With a single click on the object of interest, our model generates a score field based on feature similarities. By adjusting the binarization threshold, the segmentation can seamlessly traverse the scene hierarchy from atomic components to entire objects, and holistic portions of the scene.
Besides, users can select and segment multiple objects simultaneously through multiple clicks.
Based on the input clicks, a region-growing approach is employed to segment the mesh and extract discrete components, which can be saved as 3D assets.

\section{Experiments}

\subsection{Hierarchical 3D Segmentation}

\begin{figure*}
    \centering
    \includegraphics[width=0.9\linewidth]{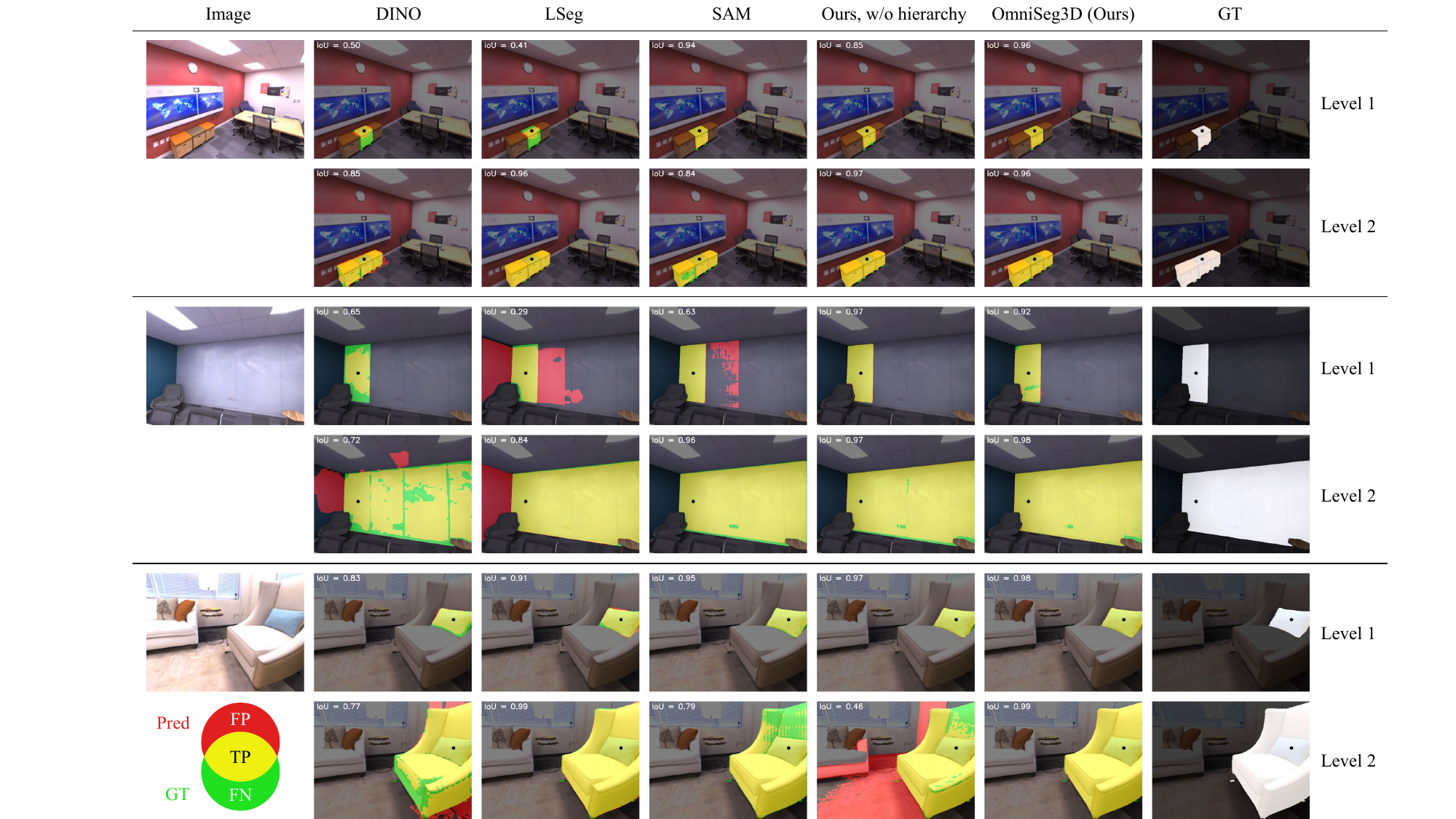}
    \caption{Comparison of hierarchical segmentation results on the Replica dataset. Prompts are shown as black dots. Colored pixels denote \textcolor{yellow}{TP: True-Positive}, \textcolor{red}{FP: False-Positive} and \textcolor{green}{FN: False-Negative} respectively.}
    \label{fig:replica}
\end{figure*}

\paragraph{Dataset.}
To quantitatively evaluate our OmniSeg3D, we set up a scene-scale dataset with hierarchical semantic annotations.
We utilize the Replica dataset~\cite{replica19arxiv} processed by Semantic-NeRF~\cite{zhi2021semanticnerf}, which comprises 8 realistic indoor scenes. We uniformly sample a total of 281 images and manually annotated each image with a query pixel $\mathbf{q}$ and two corresponding masks, the smaller one $M_{L_1}$ properly included by the larger one $M_{L_2}\supset M_{L_1}$.
$M_{L_1}$ and $M_{L_2}$ typically correspond to object parts and complete instances respectively, as shown in Fig.~\ref{fig:replica}. In case multiple levels of reasonable segmentations $M_a\subset M_b\subset M_c$ exist, we choose different pairs as the ground truth $(M_{L_1},M_{L_2})$ in different images, so that the selected masks exhibit diverse scales and represent the full range of possible hierarchical relationships present in the scene.


\paragraph{Benchmark.}
We benchmark our algorithm as follows. The model receives as input a 2D query point $\mathbf{q}$ in the given frame $\mathbf{I}$, and is expected to output a dense 2D score map $\{\mathrm{score}\,(\mathbf{p})\,|\,\mathbf{p}\in\mathbf{I}\}$.
Ideally, there exist thresholds ${th}_1>{th}_2$ which, when applied to the score map, yields $M_{L_1}\subset M_{L_2}$ respectively:
\begin{equation}
    \exists\,{th}_i\,\,s.t.\,\,M_{L_i}=\{\mathbf{p}\in\mathbf{I}\,|\,\mathrm{score}\,(\mathbf{p})>{th}_i\}.
\end{equation}
For evaluation, we choose the thresholds $({th}_1,{th}_2)$ that maximize the IoU between the predicted masks and the ground truth $(M_{L_1},M_{L_2})$, and define the metrics as:
\begin{eqnarray}
    \label{eq:iou}
    \begin{aligned}
        \mathrm{IoU}_{L_i}&=\max_{{th}_i}{\mathrm{IoU}\,(\{\mathbf{p}\in\mathbf{I}\,|\,\mathrm{score}\,(\mathbf{p})>{th}_i\},M_{L_i})}, \\
        \mathrm{IoU}_{Avg}&=(\mathrm{IoU}_{L_1}+\mathrm{IoU}_{L_2})/2.
    \end{aligned}
\end{eqnarray}

\begin{table}
    \centering
    \small
    \begin{tabular}{c|ccc}
        \toprule
        \multirow{2}*{Method} & \multicolumn{3}{c}{mIoU (\%)}\\
        & Level 1 & Level 2 & Average\\
        \midrule
        DINO \cite{caron2021dino} & 67.9 & 64.2 & 66.1\\
        LSeg \cite{li2022lseg} & 51.7 & 82.1 & 66.9\\
        SAM \cite{kirillov2023sam} & 92.8 & 80.2 & 86.5\\
        \midrule
        Ours, w/o hierarchy & \textbf{93.1} & 80.4 & 86.7\\
        OmniSeg3D (ours) & 91.3 & \textbf{88.9} & \textbf{90.1}\\
        \bottomrule
    \end{tabular}
    \caption{Comparison of hierarchical segmentation on Replica \cite{replica19arxiv}.}
    \label{tab:hierarchy}
\end{table}

\paragraph{Baseline methods.}
We first compare our OmniSeg3D with state-of-the-art 2D segmentation models and semantic feature extractors. SAM \cite{kirillov2023sam} predicts three hierarchical masks from the point query. We compare each to the ground truth masks $(M_{L_1},M_{L_2})$ and report the highest IoU. DINO \cite{caron2021dino} and LSeg \cite{li2022lseg} (based on CLIP \cite{radford2021clip}) predict a feature image, which is converted to a score map based on cosine similarities and then binarized using Eq.~\ref{eq:iou} to compute the IoU. In addition, we compare our full method with the basic implementation in Sec.~\ref{sec:basic}, i.e., 3D contrastive learning without hierarchical modelling.

\paragraph{Results.}

Tab.~\ref{tab:hierarchy} demonstrates the quantitative results of hierarchical segmentation on the Replica \cite{replica19arxiv} dataset.
Fig.~\ref{fig:replica} shows the qualitative results.
Our OmniSeg3D achieves the highest average mIoU, while substantially leading in level-2 segmentation, which features high-level semantics.

As shown in Fig.~\ref{fig:replica}, the self-supervised DINO method struggles to delineate clear object boundaries.
LSeg captures overall semantics better but fails to discriminate between instances.
SAM performs well at fine-grained segmentation, but occasionally fails to group together multiple objects or large regions, resulting in lower level-2 mIoU.
Our basic implementation without hierarchical modeling inherits these characteristics of SAM, with slightly better metrics.
Our full method degrades in level-1 segmentation due to the shifted emphasis on the omniversal task, while achieving large improvements in high-level segmentation.
This implies that the hierarchical modelling effectively aggregates fragmented part-whole correlations from multiple views.
%
We hypothesize that the 3D contrastive learning implicitly aggregates and averages the voting-based correlations from multi-view inputs, distilling a stable hierarchical semantic order into the 3D representation, thereby enhancing global-scale semantic clustering.
%

\subsection{3D Instance Segmentation}

While designed for omniversal 3D segmentation, our method is able to handle 3D instance segmentation as a sub-task.
Different from existing methods \cite{mirzaei2023spinnerf, cen2023samin3d}, OmniSeg3D does not require instance-specific training.
The 3D feature field is trained \textit{only once} for each scene and reused for different instances, while still performing competitively on datasets proposed by previous work.

We follow NVOS \cite{ren2022nvos}, SPIn-NeRF \cite{mirzaei2023spinnerf} and SA3D \cite{cen2023samin3d} to benchmark 3D instance segmentation as prompt propagation.
For each scene, given prompts (scribbles or masks) in the reference view, the algorithm is supposed to segment the instance in the target view.
The predicted mask is compared with the ground truth target view segmentation.
As shown in Tab.~\ref{tab:instance}, OmniSeg3D outperforms the baseline methods in terms of mIoU and pixel-wise classification accuracy, while alleviating the need to retrain different segmentation fields for the same scene.


\begin{table}
    \centering
    \small
    \begin{tabular}{c|c|cc}
        \toprule
        Dataset & Method & mIoU (\%) & Acc (\%)\\
        \midrule
        \multirow{4}*{NVOS} & NVOS \cite{ren2022nvos} & 70.1 & 92.0\\
        & ISRF \cite{goel2023isrf} & 83.8 & 96.4\\
        & SA3D \cite{cen2023samin3d} & 90.3 & 98.2\\
        & OmniSeg3D (ours) & \textbf{91.7} & \textbf{98.4}\\
        \midrule
        \multirow{3}*{MVSeg} & MVSeg \cite{mirzaei2023spinnerf} & 93.3 & 98.7\\
        & SA3D \cite{cen2023samin3d} & 92.8 & 98.7\\
        & OmniSeg3D (ours) & \textbf{95.2} & \textbf{99.2}\\
        \midrule
        \multirow{3}*{Replica} & MVSeg \cite{mirzaei2023spinnerf} & 32.4 & -\\
        & SA3D \cite{cen2023samin3d} & 83.0 & -\\
        & OmniSeg3D (ours) & \textbf{84.4} & -\\
        \bottomrule
    \end{tabular}
    \caption{Quantitative comparison of instance segmentation.}
    \label{tab:instance}
\end{table}


\subsection{Ablation Studies}
\label{sec:ablation}

\paragraph{Hierarchical decay.}
As illustrated in Eq.~\ref{eq:reg}, we apply a decay $\lambda\in[0,1]$ to downweight the contrastive loss for patches of lower correlation with the anchor.
Setting $\lambda=0$ resembles the basic implementation without hierarchical modeling, while setting $\lambda=1$ puts equal emphasis on samples from all hierarchies, enhancing high-level semantics.
Tab.~\ref{tab:ablation} demonstrates hierarchical segmentation results on the Replica dataset.
With the increase of $\lambda$, $\mathrm{IoU}_{L_1}$ decreases while $\mathrm{IoU}_{L_2}$ increases, reaching $\mathrm{IoU}_{L_1}\approx\mathrm{IoU}_{L_2}$ at $\lambda=1$.
We choose $\lambda=0.5$ with the highest average mIoU, implying a balance between local and global semantic clustering.
For instance segmentation, the influence of $\lambda$ on mIoU is counteracted when averaged on instances with various sizes and containing different levels of hierarchies.

\begin{table}
    \centering
    \small
    \begin{tabular}{cc|ccc|c}
        \toprule
        Hierar. & Per-level & \multicolumn{3}{c|}{Hierar. mIoU (\%)} & Instance\\
        model & decay $\lambda$ & Lv.1 & Lv.2 & Avg. & mIoU (\%)\\
        \midrule
        $\times$ & - & \textbf{93.1} & 80.4 & 86.7 & 83.6\\
        \checkmark & 0.1 & 92.5 & 84.7 & 88.6 & 84.3\\
        \checkmark & 0.2 & 92.1 & 86.5 & 89.4 & \textbf{84.6}\\
        \checkmark & 0.5 & 91.3 & 88.9 & \textbf{90.1} & 84.4\\
        \checkmark & 1 & 89.2 & \textbf{89.2} & 89.2 & 83.3\\
        \bottomrule
    \end{tabular}
    \caption{Ablation of hierarchical modelling on Replica.}
    \label{tab:ablation}
\end{table}

\paragraph{Feature dimension.}
We study how the dimension $D$ of semantic features affects the performance of hierarchical contrastive clustering. As shown in Tab.~\ref{tab:dim}, the average mIoU first increases with $D$, then nearly saturates beyond $D=16$. Therefore, we assume $D=16$ is sufficient for our algorithm.

\begin{table}
    \centering
    \small
    \begin{tabular}{c|cccccc}
        \toprule
        Feat. dim. & 4 & 8 & 16 & 32 & 64 & 128\\
        \midrule
        Avg. mIoU & 89.8 & 91.8 & 93.0 & 93.0 & 93.1 & 93.2\\
        \bottomrule
    \end{tabular}
    \caption{Ablation of feature dimensions on \texttt{room-0} of Replica.}
    \label{tab:dim}
\end{table}

\section{Limitations}


\begin{figure}[t] 
\centering
    \includegraphics[width=1.0\linewidth]{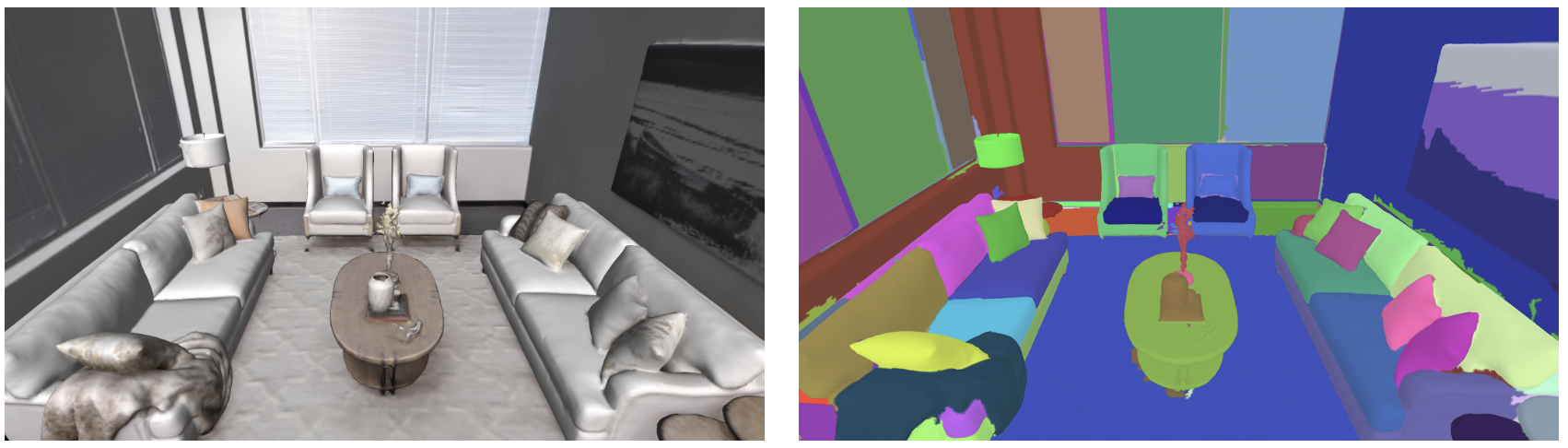} 
    \caption{Scene discretization by feature clustering on mesh automatically without click.}\label{fig:limitation}
\end{figure}


Due to the absence of a clear definition for hierarchy levels, there is no assurance that the objects will be segmented at the same level by simply clustering features (see Fig.~\ref{fig:limitation}). To address this issue, text-aligned hierarchical segmentation may be a future direction.
Besides, since the contrastive learning is applied on single images, two objects that have never appeared in the same image may have similar semantic feature. This problem can be alleviated by introducing local geometric continuity, but global contrastive learning across images is also a topic worth exploring.

\section{Conclusion}
In this paper, we propose OmniSeg3D, an omniversal segmentation method that facilitates holistic understanding of 3D scenes. 
Leveraging a hierarchical representation and a hierarchical contrastive learning framework, OmniSeg3D effectively transforms inconsistent 2D segmentations into a globally consistent 3D feature field while retaining hierarchical information, which enables correct hierarchical 3D sensing and high-quality object segmentation performance. Besides, variant interactive functionalities including hierarchical inference, multi-object selection, and global discretization are realized, which may further enable downstream applications in the field of 3D data annotation, robotics and virtual reality.

{
    \small
    \bibliographystyle{ieeenat_fullname}
    \bibliography{main}

\begin{thebibliography}{53}
\providecommand{\natexlab}[1]{#1}
\providecommand{\url}[1]{\texttt{#1}}
\expandafter\ifx\csname urlstyle\endcsname\relax
  \providecommand{\doi}[1]{doi: #1}\else
  \providecommand{\doi}{doi: \begingroup \urlstyle{rm}\Url}\fi

\bibitem[Achanta et~al.(2012)Achanta, Shaji, Smith, Lucchi, Fua, and S{\"u}sstrunk]{achanta2012slic}
Radhakrishna Achanta, Appu Shaji, Kevin Smith, Aurelien Lucchi, Pascal Fua, and Sabine S{\"u}sstrunk.
\newblock Slic superpixels compared to state-of-the-art superpixel methods.
\newblock \emph{IEEE transactions on pattern analysis and machine intelligence}, 34\penalty0 (11):\penalty0 2274--2282, 2012.

\bibitem[Bhalgat et~al.(2023)Bhalgat, Laina, Henriques, Zisserman, and Vedaldi]{bhalgat2023contrastivelift}
Yash Bhalgat, Iro Laina, Jo{\~a}o~F Henriques, Andrew Zisserman, and Andrea Vedaldi.
\newblock Contrastive lift: 3d object instance segmentation by slow-fast contrastive fusion.
\newblock \emph{arXiv preprint arXiv:2306.04633}, 2023.

\bibitem[Bing et~al.(2022)Bing, Chen, and Yang]{bing2022dmnerf}
WANG Bing, Lu Chen, and Bo Yang.
\newblock Dm-nerf: 3d scene geometry decomposition and manipulation from 2d images.
\newblock In \emph{The Eleventh International Conference on Learning Representations}, 2022.

\bibitem[Caron et~al.(2021)Caron, Touvron, Misra, J\'egou, Mairal, Bojanowski, and Joulin]{caron2021dino}
Mathilde Caron, Hugo Touvron, Ishan Misra, Herv\'e J\'egou, Julien Mairal, Piotr Bojanowski, and Armand Joulin.
\newblock Emerging properties in self-supervised vision transformers.
\newblock In \emph{Proceedings of the International Conference on Computer Vision (ICCV)}, 2021.

\bibitem[Cen et~al.(2023)Cen, Zhou, Fang, Shen, Xie, Zhang, and Tian]{cen2023samin3d}
Jiazhong Cen, Zanwei Zhou, Jiemin Fang, Wei Shen, Lingxi Xie, Xiaopeng Zhang, and Qi Tian.
\newblock Segment anything in 3d with nerfs.
\newblock \emph{arXiv preprint arXiv:2304.12308}, 2023.

\bibitem[Chen et~al.(2017)Chen, Papandreou, Kokkinos, Murphy, and Yuille]{chen2017deeplab}
Liang-Chieh Chen, George Papandreou, Iasonas Kokkinos, Kevin Murphy, and Alan~L Yuille.
\newblock Deeplab: Semantic image segmentation with deep convolutional nets, atrous convolution, and fully connected crfs.
\newblock \emph{IEEE transactions on pattern analysis and machine intelligence}, 40\penalty0 (4):\penalty0 834--848, 2017.

\bibitem[Chen et~al.(2022)Chen, Zhao, Zhang, Duan, Qi, and Zhao]{chen2022focalclick}
Xi Chen, Zhiyan Zhao, Yilei Zhang, Manni Duan, Donglian Qi, and Hengshuang Zhao.
\newblock Focalclick: Towards practical interactive image segmentation.
\newblock In \emph{Proceedings of the IEEE/CVF Conference on Computer Vision and Pattern Recognition}, pages 1300--1309, 2022.

\bibitem[Chen et~al.(2020)Chen, Tagliasacchi, and Zhang]{chen2020bspnet}
Zhiqin Chen, Andrea Tagliasacchi, and Hao Zhang.
\newblock Bsp-net: Generating compact meshes via binary space partitioning.
\newblock In \emph{Proceedings of the IEEE/CVF Conference on Computer Vision and Pattern Recognition}, pages 45--54, 2020.

\bibitem[Cheng et~al.(2021)Cheng, Schwing, and Kirillov]{cheng2021perpixel}
Bowen Cheng, Alex Schwing, and Alexander Kirillov.
\newblock Per-pixel classification is not all you need for semantic segmentation.
\newblock \emph{Advances in Neural Information Processing Systems}, 34:\penalty0 17864--17875, 2021.

\bibitem[Cheng et~al.(2022)Cheng, Misra, Schwing, Kirillov, and Girdhar]{cheng2022maskedattention}
Bowen Cheng, Ishan Misra, Alexander~G Schwing, Alexander Kirillov, and Rohit Girdhar.
\newblock Masked-attention mask transformer for universal image segmentation.
\newblock In \emph{Proceedings of the IEEE/CVF conference on computer vision and pattern recognition}, pages 1290--1299, 2022.

\bibitem[Coleman and Andrews(1979)]{coleman1979image}
Guy~Barrett Coleman and Harry~C Andrews.
\newblock Image segmentation by clustering.
\newblock \emph{Proceedings of the IEEE}, 67\penalty0 (5):\penalty0 773--785, 1979.

\bibitem[Dorninger and Nothegger(2007)]{dorninger20073d}
Peter Dorninger and Clemens Nothegger.
\newblock 3d segmentation of unstructured point clouds for building modelling.
\newblock \emph{International Archives of the Photogrammetry, Remote Sensing and Spatial Information Sciences}, 35\penalty0 (3/W49A):\penalty0 191--196, 2007.

\bibitem[Felzenszwalb and Huttenlocher(2004)]{felzenszwalb2004efficient}
Pedro~F Felzenszwalb and Daniel~P Huttenlocher.
\newblock Efficient graph-based image segmentation.
\newblock \emph{International journal of computer vision}, 59:\penalty0 167--181, 2004.

\bibitem[Goel et~al.(2023)Goel, Sirikonda, Saini, and Narayanan]{goel2023isrf}
Rahul Goel, Dhawal Sirikonda, Saurabh Saini, and PJ Narayanan.
\newblock Interactive segmentation of radiance fields.
\newblock In \emph{Proceedings of the IEEE/CVF Conference on Computer Vision and Pattern Recognition}, pages 4201--4211, 2023.

\bibitem[Han et~al.(2020)Han, Zheng, Xu, and Fang]{han2020occuseg}
Lei Han, Tian Zheng, Lan Xu, and Lu Fang.
\newblock Occuseg: Occupancy-aware 3d instance segmentation.
\newblock In \emph{Proceedings of the IEEE/CVF conference on computer vision and pattern recognition}, pages 2940--2949, 2020.

\bibitem[He et~al.(2017)He, Gkioxari, Doll{\'a}r, and Girshick]{he2017maskrcnn}
Kaiming He, Georgia Gkioxari, Piotr Doll{\'a}r, and Ross Girshick.
\newblock Mask r-cnn.
\newblock In \emph{Proceedings of the IEEE international conference on computer vision}, pages 2961--2969, 2017.

\bibitem[H{\"o}hne and Hanson(1992)]{hohne1992trad3Dseg1}
Karl~Heinz H{\"o}hne and William~A Hanson.
\newblock Interactive 3d segmentation of mri and ct volumes using morphological operations.
\newblock \emph{Journal of computer assisted tomography}, 16\penalty0 (2):\penalty0 285--294, 1992.

\bibitem[Hu et~al.(2020)Hu, Yang, Xie, Rosa, Guo, Wang, Trigoni, and Markham]{hu2020randlanet}
Qingyong Hu, Bo Yang, Linhai Xie, Stefano Rosa, Yulan Guo, Zhihua Wang, Niki Trigoni, and Andrew Markham.
\newblock Randla-net: Efficient semantic segmentation of large-scale point clouds.
\newblock In \emph{Proceedings of the IEEE/CVF conference on computer vision and pattern recognition}, pages 11108--11117, 2020.

\bibitem[Huang and You(2016)]{huang2016point}
Jing Huang and Suya You.
\newblock Point cloud labeling using 3d convolutional neural network.
\newblock In \emph{2016 23rd International Conference on Pattern Recognition (ICPR)}, pages 2670--2675. IEEE, 2016.

\bibitem[Kerr et~al.(2023)Kerr, Kim, Goldberg, Kanazawa, and Tancik]{kerr2023lerf}
Justin Kerr, Chung~Min Kim, Ken Goldberg, Angjoo Kanazawa, and Matthew Tancik.
\newblock Lerf: Language embedded radiance fields.
\newblock In \emph{Proceedings of the IEEE/CVF International Conference on Computer Vision}, pages 19729--19739, 2023.

\bibitem[Kirillov et~al.(2019)Kirillov, He, Girshick, Rother, and Doll{\'a}r]{kirillov2019panopticseg}
Alexander Kirillov, Kaiming He, Ross Girshick, Carsten Rother, and Piotr Doll{\'a}r.
\newblock Panoptic segmentation.
\newblock In \emph{Proceedings of the IEEE/CVF conference on computer vision and pattern recognition}, pages 9404--9413, 2019.

\bibitem[Kirillov et~al.(2023)Kirillov, Mintun, Ravi, Mao, Rolland, Gustafson, Xiao, Whitehead, Berg, Lo, et~al.]{kirillov2023sam}
Alexander Kirillov, Eric Mintun, Nikhila Ravi, Hanzi Mao, Chloe Rolland, Laura Gustafson, Tete Xiao, Spencer Whitehead, Alexander~C Berg, Wan-Yen Lo, et~al.
\newblock Segment anything.
\newblock \emph{arXiv preprint arXiv:2304.02643}, 2023.

\bibitem[Kobayashi et~al.(2022)Kobayashi, Matsumoto, and Sitzmann]{kobayashi2022dff}
Sosuke Kobayashi, Eiichi Matsumoto, and Vincent Sitzmann.
\newblock Decomposing nerf for editing via feature field distillation.
\newblock \emph{Advances in Neural Information Processing Systems}, 35:\penalty0 23311--23330, 2022.

\bibitem[Li et~al.(2022)Li, Weinberger, Belongie, Koltun, and Ranftl]{li2022lseg}
Boyi Li, Kilian~Q Weinberger, Serge Belongie, Vladlen Koltun, and Rene Ranftl.
\newblock Language-driven semantic segmentation.
\newblock In \emph{International Conference on Learning Representations}, 2022.

\bibitem[Li et~al.(2020)Li, Zhou, Xiong, and Hoi]{li2020pcl}
Junnan Li, Pan Zhou, Caiming Xiong, and Steven Hoi.
\newblock Prototypical contrastive learning of unsupervised representations.
\newblock In \emph{International Conference on Learning Representations}, 2020.

\bibitem[Liu et~al.(2022)Liu, Zheng, Lin, Ni, and Fang]{liu2022insconv}
Leyao Liu, Tian Zheng, Yun-Jou Lin, Kai Ni, and Lu Fang.
\newblock Ins-conv: Incremental sparse convolution for online 3d segmentation.
\newblock In \emph{Proceedings of the IEEE/CVF Conference on Computer Vision and Pattern Recognition}, pages 18975--18984, 2022.

\bibitem[Liu et~al.(2023)Liu, Xu, Bertasius, and Niethammer]{liu2023simpleclick}
Qin Liu, Zhenlin Xu, Gedas Bertasius, and Marc Niethammer.
\newblock Simpleclick: Interactive image segmentation with simple vision transformers.
\newblock In \emph{Proceedings of the IEEE/CVF International Conference on Computer Vision}, pages 22290--22300, 2023.

\bibitem[Liu et~al.(2019)Liu, Tang, Lin, and Han]{liu2019point}
Zhijian Liu, Haotian Tang, Yujun Lin, and Song Han.
\newblock Point-voxel cnn for efficient 3d deep learning.
\newblock \emph{Advances in Neural Information Processing Systems}, 32, 2019.

\bibitem[Long et~al.(2015)Long, Shelhamer, and Darrell]{long2015fcn}
Jonathan Long, Evan Shelhamer, and Trevor Darrell.
\newblock Fully convolutional networks for semantic segmentation.
\newblock In \emph{Proceedings of the IEEE conference on computer vision and pattern recognition}, pages 3431--3440, 2015.

\bibitem[Mildenhall et~al.(2021)Mildenhall, Srinivasan, Tancik, Barron, Ramamoorthi, and Ng]{mildenhall2021nerf}
Ben Mildenhall, Pratul~P Srinivasan, Matthew Tancik, Jonathan~T Barron, Ravi Ramamoorthi, and Ren Ng.
\newblock Nerf: Representing scenes as neural radiance fields for view synthesis.
\newblock \emph{Communications of the ACM}, 65\penalty0 (1):\penalty0 99--106, 2021.

\bibitem[Mirzaei et~al.(2023)Mirzaei, Aumentado-Armstrong, Derpanis, Kelly, Brubaker, Gilitschenski, and Levinshtein]{mirzaei2023spinnerf}
Ashkan Mirzaei, Tristan Aumentado-Armstrong, Konstantinos~G Derpanis, Jonathan Kelly, Marcus~A Brubaker, Igor Gilitschenski, and Alex Levinshtein.
\newblock Spin-nerf: Multiview segmentation and perceptual inpainting with neural radiance fields.
\newblock In \emph{Proceedings of the IEEE/CVF Conference on Computer Vision and Pattern Recognition}, pages 20669--20679, 2023.

\bibitem[Mo et~al.(2019{\natexlab{a}})Mo, Guerrero, Yi, Su, Wonka, Mitra, and Guibas]{mo2019structurenet}
Kaichun Mo, Paul Guerrero, Li Yi, Hao Su, Peter Wonka, Niloy Mitra, and Leonidas~J Guibas.
\newblock Structurenet: Hierarchical graph networks for 3d shape generation.
\newblock \emph{arXiv preprint arXiv:1908.00575}, 2019{\natexlab{a}}.

\bibitem[Mo et~al.(2019{\natexlab{b}})Mo, Zhu, Chang, Yi, Tripathi, Guibas, and Su]{mo2019partnet}
Kaichun Mo, Shilin Zhu, Angel~X Chang, Li Yi, Subarna Tripathi, Leonidas~J Guibas, and Hao Su.
\newblock Partnet: A large-scale benchmark for fine-grained and hierarchical part-level 3d object understanding.
\newblock In \emph{Proceedings of the IEEE/CVF conference on computer vision and pattern recognition}, pages 909--918, 2019{\natexlab{b}}.

\bibitem[M{\"u}ller et~al.(2022)M{\"u}ller, Evans, Schied, and Keller]{muller2022instantngp}
Thomas M{\"u}ller, Alex Evans, Christoph Schied, and Alexander Keller.
\newblock Instant neural graphics primitives with a multiresolution hash encoding.
\newblock \emph{ACM Transactions on Graphics (ToG)}, 41\penalty0 (4):\penalty0 1--15, 2022.

\bibitem[Peng et~al.(2023)Peng, Genova, Jiang, Tagliasacchi, Pollefeys, Funkhouser, et~al.]{peng2023openscene}
Songyou Peng, Kyle Genova, Chiyu Jiang, Andrea Tagliasacchi, Marc Pollefeys, Thomas Funkhouser, et~al.
\newblock Openscene: 3d scene understanding with open vocabularies.
\newblock In \emph{Proceedings of the IEEE/CVF Conference on Computer Vision and Pattern Recognition}, pages 815--824, 2023.

\bibitem[Radford et~al.(2021)Radford, Kim, Hallacy, Ramesh, Goh, Agarwal, Sastry, Askell, Mishkin, Clark, et~al.]{radford2021clip}
Alec Radford, Jong~Wook Kim, Chris Hallacy, Aditya Ramesh, Gabriel Goh, Sandhini Agarwal, Girish Sastry, Amanda Askell, Pamela Mishkin, Jack Clark, et~al.
\newblock Learning transferable visual models from natural language supervision.
\newblock In \emph{International conference on machine learning}, pages 8748--8763. PMLR, 2021.

\bibitem[Ren et~al.(2022)Ren, Agarwala, Russell, Schwing, and Wang]{ren2022nvos}
Zhongzheng Ren, Aseem Agarwala, Bryan Russell, Alexander~G Schwing, and Oliver Wang.
\newblock Neural volumetric object selection.
\newblock In \emph{Proceedings of the IEEE/CVF Conference on Computer Vision and Pattern Recognition}, pages 6133--6142, 2022.

\bibitem[Schnabel et~al.(2007)Schnabel, Wahl, and Klein]{schnabel2007efficientransac}
Ruwen Schnabel, Roland Wahl, and Reinhard Klein.
\newblock Efficient ransac for point-cloud shape detection.
\newblock In \emph{Computer graphics forum}, pages 214--226. Wiley Online Library, 2007.

\bibitem[Siddiqui et~al.(2023)Siddiqui, Porzi, Bul{\`o}, M{\"u}ller, Nie{\ss}ner, Dai, and Kontschieder]{siddiqui2023panopticlift}
Yawar Siddiqui, Lorenzo Porzi, Samuel~Rota Bul{\`o}, Norman M{\"u}ller, Matthias Nie{\ss}ner, Angela Dai, and Peter Kontschieder.
\newblock Panoptic lifting for 3d scene understanding with neural fields.
\newblock In \emph{Proceedings of the IEEE/CVF Conference on Computer Vision and Pattern Recognition}, pages 9043--9052, 2023.

\bibitem[Sofiiuk et~al.(2022)Sofiiuk, Petrov, and Konushin]{sofiiuk2022ritm}
Konstantin Sofiiuk, Ilya~A Petrov, and Anton Konushin.
\newblock Reviving iterative training with mask guidance for interactive segmentation.
\newblock In \emph{2022 IEEE International Conference on Image Processing (ICIP)}, pages 3141--3145. IEEE, 2022.

\bibitem[Straub et~al.(2019)Straub, Whelan, Ma, Chen, Wijmans, Green, Engel, Mur-Artal, Ren, Verma, Clarkson, Yan, Budge, Yan, Pan, Yon, Zou, Leon, Carter, Briales, Gillingham, Mueggler, Pesqueira, Savva, Batra, Strasdat, Nardi, Goesele, Lovegrove, and Newcombe]{replica19arxiv}
Julian Straub, Thomas Whelan, Lingni Ma, Yufan Chen, Erik Wijmans, Simon Green, Jakob~J. Engel, Raul Mur-Artal, Carl Ren, Shobhit Verma, Anton Clarkson, Mingfei Yan, Brian Budge, Yajie Yan, Xiaqing Pan, June Yon, Yuyang Zou, Kimberly Leon, Nigel Carter, Jesus Briales, Tyler Gillingham, Elias Mueggler, Luis Pesqueira, Manolis Savva, Dhruv Batra, Hauke~M. Strasdat, Renzo~De Nardi, Michael Goesele, Steven Lovegrove, and Richard Newcombe.
\newblock The {R}eplica dataset: A digital replica of indoor spaces.
\newblock \emph{arXiv preprint arXiv:1906.05797}, 2019.

\bibitem[Takmaz et~al.(2023)Takmaz, Fedele, Sumner, Pollefeys, Tombari, and Engelmann]{takmaz2023openmask3d}
Ay{\c{c}}a Takmaz, Elisabetta Fedele, Robert~W Sumner, Marc Pollefeys, Federico Tombari, and Francis Engelmann.
\newblock Openmask3d: Open-vocabulary 3d instance segmentation.
\newblock \emph{arXiv preprint arXiv:2306.13631}, 2023.

\bibitem[Vaswani et~al.(2017)Vaswani, Shazeer, Parmar, Uszkoreit, Jones, Gomez, Kaiser, and Polosukhin]{vaswani2017attention}
Ashish Vaswani, Noam Shazeer, Niki Parmar, Jakob Uszkoreit, Llion Jones, Aidan~N Gomez, {\L}ukasz Kaiser, and Illia Polosukhin.
\newblock Attention is all you need.
\newblock \emph{Advances in neural information processing systems}, 30, 2017.

\bibitem[Wang and Neumann(2018)]{wang2018depth}
Weiyue Wang and Ulrich Neumann.
\newblock Depth-aware cnn for rgb-d segmentation.
\newblock In \emph{Proceedings of the European conference on computer vision (ECCV)}, pages 135--150, 2018.

\bibitem[Wu et~al.(2022)Wu, Liu, Chen, Li, Zheng, Cai, and Zheng]{wu2022objectsdf}
Qianyi Wu, Xian Liu, Yuedong Chen, Kejie Li, Chuanxia Zheng, Jianfei Cai, and Jianmin Zheng.
\newblock Object-compositional neural implicit surfaces.
\newblock In \emph{European Conference on Computer Vision}, pages 197--213. Springer, 2022.

\bibitem[Xing et~al.(2020)Xing, Wang, and Zeng]{xing2020malleable}
Yajie Xing, Jingbo Wang, and Gang Zeng.
\newblock Malleable 2.5 d convolution: Learning receptive fields along the depth-axis for rgb-d scene parsing.
\newblock In \emph{European Conference on Computer Vision}, pages 555--571. Springer, 2020.

\bibitem[Yang et~al.(2019)Yang, Wang, Clark, Hu, Wang, Markham, and Trigoni]{yang20193dbonet}
Bo Yang, Jianan Wang, Ronald Clark, Qingyong Hu, Sen Wang, Andrew Markham, and Niki Trigoni.
\newblock Learning object bounding boxes for 3d instance segmentation on point clouds.
\newblock \emph{Advances in neural information processing systems}, 32, 2019.

\bibitem[Yi et~al.(2019)Yi, Zhao, Wang, Sung, and Guibas]{yi2019gspn}
Li Yi, Wang Zhao, He Wang, Minhyuk Sung, and Leonidas~J Guibas.
\newblock Gspn: Generative shape proposal network for 3d instance segmentation in point cloud.
\newblock In \emph{Proceedings of the IEEE/CVF Conference on Computer Vision and Pattern Recognition}, pages 3947--3956, 2019.

\bibitem[Yu et~al.(2022)Yu, Chen, Li, Sanghi, Shayani, Mahdavi-Amiri, and Zhang]{yu2022caprinet}
Fenggen Yu, Zhiqin Chen, Manyi Li, Aditya Sanghi, Hooman Shayani, Ali Mahdavi-Amiri, and Hao Zhang.
\newblock Capri-net: Learning compact cad shapes with adaptive primitive assembly.
\newblock In \emph{Proceedings of the IEEE/CVF Conference on Computer Vision and Pattern Recognition}, pages 11768--11778, 2022.

\bibitem[Zhang et~al.(2022)Zhang, Xu, Xiong, and Ramaiah]{zhang2022useallthelabels}
Shu Zhang, Ran Xu, Caiming Xiong, and Chetan Ramaiah.
\newblock Use all the labels: A hierarchical multi-label contrastive learning framework.
\newblock In \emph{Proceedings of the IEEE/CVF Conference on Computer Vision and Pattern Recognition}, pages 16660--16669, 2022.

\bibitem[Zhao et~al.(2017)Zhao, Shi, Qi, Wang, and Jia]{zhao2017pyramid}
Hengshuang Zhao, Jianping Shi, Xiaojuan Qi, Xiaogang Wang, and Jiaya Jia.
\newblock Pyramid scene parsing network.
\newblock In \emph{Proceedings of the IEEE conference on computer vision and pattern recognition}, pages 2881--2890, 2017.

\bibitem[Zheng et~al.(2021)Zheng, Lu, Zhao, Zhu, Luo, Wang, Fu, Feng, Xiang, Torr, et~al.]{zheng2021rethinking}
Sixiao Zheng, Jiachen Lu, Hengshuang Zhao, Xiatian Zhu, Zekun Luo, Yabiao Wang, Yanwei Fu, Jianfeng Feng, Tao Xiang, Philip~HS Torr, et~al.
\newblock Rethinking semantic segmentation from a sequence-to-sequence perspective with transformers.
\newblock In \emph{Proceedings of the IEEE/CVF conference on computer vision and pattern recognition}, pages 6881--6890, 2021.

\bibitem[Zhi et~al.(2021)Zhi, Laidlow, Leutenegger, and Davison]{zhi2021semanticnerf}
Shuaifeng Zhi, Tristan Laidlow, Stefan Leutenegger, and Andrew~J Davison.
\newblock In-place scene labelling and understanding with implicit scene representation.
\newblock In \emph{Proceedings of the IEEE/CVF International Conference on Computer Vision}, pages 15838--15847, 2021.

\end{thebibliography}
}


\end{document}